\documentclass[10pt, twocolumn, letterpaper]{article}

\usepackage{iccv}
\usepackage{times}
\usepackage{epsfig}
\usepackage{graphicx}
\usepackage{amsmath}
\usepackage{amssymb}
\usepackage[colorlinks, linkcolor=red]{hyperref}
\usepackage{url}

\iccvfinalcopy

\def\iccvPaperID{****}

\begin{document}

\title{A Computing Kernel for Network Binarization on PyTorch}

\author{Xianda Xu\\
University of Electronic Science and Technology of China\\
{\small Chengdu, China}\\
{\tt\small xiandaxu@std.uestc.edu.cn}
\and
Marco Pedersoli\\
Ecole de Technologie Supérieure\\
{\small Montreal, Canada}\\
{\tt\small Marco.Pedersoli@etsmtl.ca}
}

\maketitle

\begin{abstract}
	Deep Neural Networks have now achieved state-of-the-art results in a wide range of tasks including image classification, object detection and so on. However, they are both computation consuming and memory intensive, making them difficult to deploy on low-power devices. Network binarization is one of the existing effective techniques for model compression and acceleration, but there is no computing kernel yet to support it on PyTorch. In this paper we developed a computing kernel supporting 1-bit xnor and bitcount computation on PyTorch. Experimental results show that our kernel could accelerate the inference of the binarized neural network by 3 times in GPU and by 4.5 times in CPU compared with the control group. \footnote{Our code is now available at \url{https://github.com/brycexu/BNN_Kernel}.}
	
\end{abstract}

\section{Introduction}
	Today, deep neural networks have achieved remarkable results in real-world tasks such as image classification \cite{he2016deep}\cite{krizhevsky2012imagenet}, object detection \cite{ren2015faster}\cite{zhang2016joint}, semantic segmentation \cite{girshick2014rich}\cite{long2015fully} and so on. However, energy efficiency has become the bottleneck for deploying deep neural networks on small devices like mobile phones and embedded devices since they are area-and-battery constrained.
	
	Current deep neural networks consist of hundreds of millions of parameters, which brings about two following issues. (1) Energy consumption. One hidden layer can be extracted as {\small $A=\sigma(X\cdot W^T+B)$} where {\small$X$} denotes the input feature, {\small$W$} denotes the parameter matrix, {\small$B$} denotes the offset matrix and {\small$\sigma$} denotes the activation function. We notice that the {\small $Gemm$} (general matrix to matrix multiplication) operation takes most computation time in both training and testing. (2) Memory cost. Normally, parameters in the deep neural network are stored in the format of float-32. AlexNet \cite{krizhevsky2012imagenet} achieved a remarkable performance but with 60 million parameters, it takes totally 240 MB for storage.
	
	Network binarization is one of the existing techniques for model compression and acceleration. It quantifies the model to binary numbers. In \cite{courbariaux2016binarized}, both weights and activations are binarized and the model could achieve a 89 \% accuracy on {\small CIFAR-10}.
	
	Unfortunately, at the time of this writing, {\small PyTorch}  \cite{paszke2017automatic} does not support 1-bit bitwidth operations and available binarized neural networks released using {\small PyTorch} by now are more of a simulation because they still use 32-bit floating point representation for the tensors and still use the {\small $Gemm$}-{\small $Accumulation$} operation in the convolutional layer. There is no actual acceleration and compression in these models.
	
	It inspired us to develop a computing kernel for network binarization on {\small PyTorch}. Firstly, it should be able to encode tensors from 32-bit floating point representation to 1-bit representation. Secondly, it should replace the {\small $Gemm$}-{\small $Accumulation$} operation with the {\small $Xnor$}-{\small $Bitcount$} operation.
	
	Contributions we have made in our work include:
	\begin{itemize}
		\item[$\cdot$]	We build a computing kernel using {\small C} and {\small CUDA}. It supports {\small $Xnor$}-{\small $Bitcount$} operations on the encoded input and weight to get the output result.
		\item[$\cdot$] Our computing kernel can be wrapped into a {\small Python} loadable shared library thus it can be used on {\small PyTorch}.
		\item[$\cdot$] Experimental results show that our kernel could accelerate the inference of a binarized neural network by 3 times in GPU and 4.5 times in CPU compared to the control group. 
	\end{itemize}

	The rest of the paper is organized as follows: In section \ref{label:cp}, we talk about how convolution is implemented on {\small PyTorch}. In section \ref{label:m}, we propose our computing kernel and explain how it works. In section \ref{label:e}, experimental results on GPU and CPU are presented. We draw our conclusions in section \ref{label:c} and make a discussion in section \ref{label:d}.
	
	\begin{figure*}[htbp]
		\centering
		\includegraphics[height=3.7cm,width=14cm]{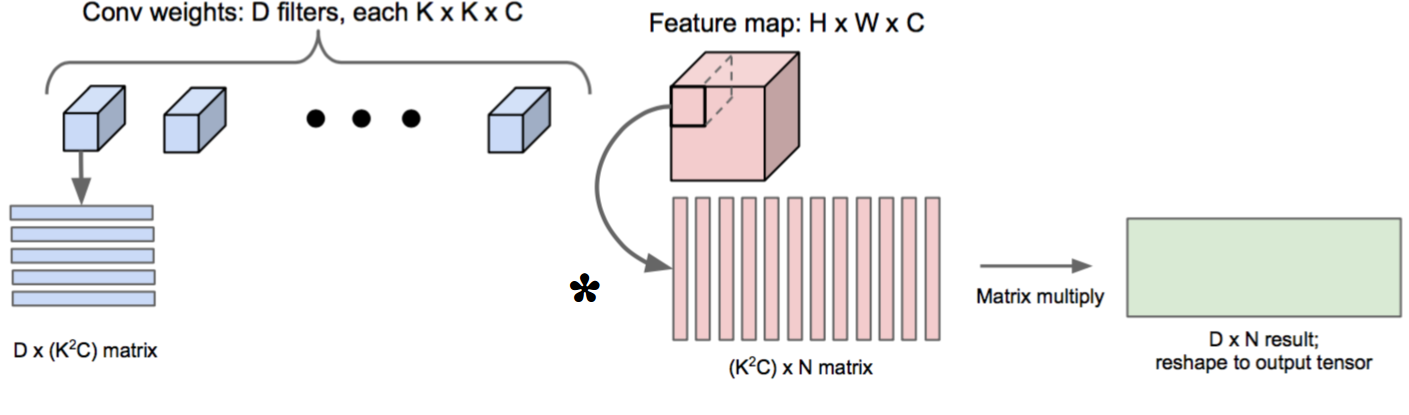}
		\caption{The $im2col$ Operation. The input feature map is {\small $[H,W,C,1]$} and is converted into a {\small $[K^2C,N]$} matrix. The filter matrix is {\small $[D,K,K,C]$} and is converted into a {\small $[D,K^2C]$} matrix. It enables the {\small $Gemm$}-{\small $Accumulation$} operation in the convolutional layer.}
		\label{Figure 1}
	\end{figure*}

\section{Convolution in Practice}
\label{label:cp}
	In this section, we would like to talk about how convolution is normally implemented in practice like on {\small PyTorch}. This may help understand what  is changed in our computing kernel.

\subsection{Im2col}
	
	For a convolutional layer, let $X$ be the input feature map, $W$ be the filter matrix and $A$ be the output feature map. $kH$ and $kW$ represent the kernel height and the kernel width. $C$ and $D$ represent the input channel and the output channel.
	
	The basic convolution operation to get each element $a_{i,j,d}$ in $A$ is performed as:
	$$
	a_{i,j,d} = \sum^{kH-1}_{h=0}\sum^{kW-1}_{w=0}\sum_{c}w_{d,h,w,c}\cdot x_{i+h,j+w,c}
	$$ 
	
	We can see that this method is quite time-consuming. So, in practice, we convert the basic convolution operation to the {\small $Gemm$}-{\small $Accumulation$} operation (general matrix to matrix multiplication) by using an operation named $im2col$ which arranges the data in a way that the output can be achieved by a simple matrix multiplication as shown in Figure \ref{Figure 1}.
	
	Notice that the result here is a {\small $[D,N,1]$} matrix so we also need to reshape it to the output feature map. This process is the inverse process of $im2col$ which is called $col2im$.
	
\subsection{Forward Graph}
	In this work, we only consider the acceleration in the inference of the binarized neural network. The forward graph used in {\small PyTorch} is shown in Figure \ref{Figure 2}.

	Both the weight matrix and the input feature map are converted by the $im2col$ operation into matrixes which are then used to perform the {\small $Gemm$}-{\small $Accumulation$} operation. The bias matrix also needs to be reshaped to perform the $addmm$ operation with the output of the {\small $Gemm$}-{\small $Accumulation$} operation. The result shoud be converted by the $col2im$ operation before we get the output feature map.

\section{Methology}
\label{label:m}

\subsection{Encoding}
	
	{\small PyTorch} normally uses the 32-bit floating points representation for the tensors, which is called {\small $FloatTensor$}. Available binarized neural networks implemented with {\small PyTorch} are more of a simulation because they are still using float-32 tensors in computation although they quantify these tensors to binary numbers with the {\small $Sign$} function. So, the first thing we need to if we want to implement a "real" binarized neural network with {\small PyTorch} is to encode the tensors.
	
	\begin{figure}[!htbp]
		\centering
		\includegraphics[height=10cm,width=6cm]{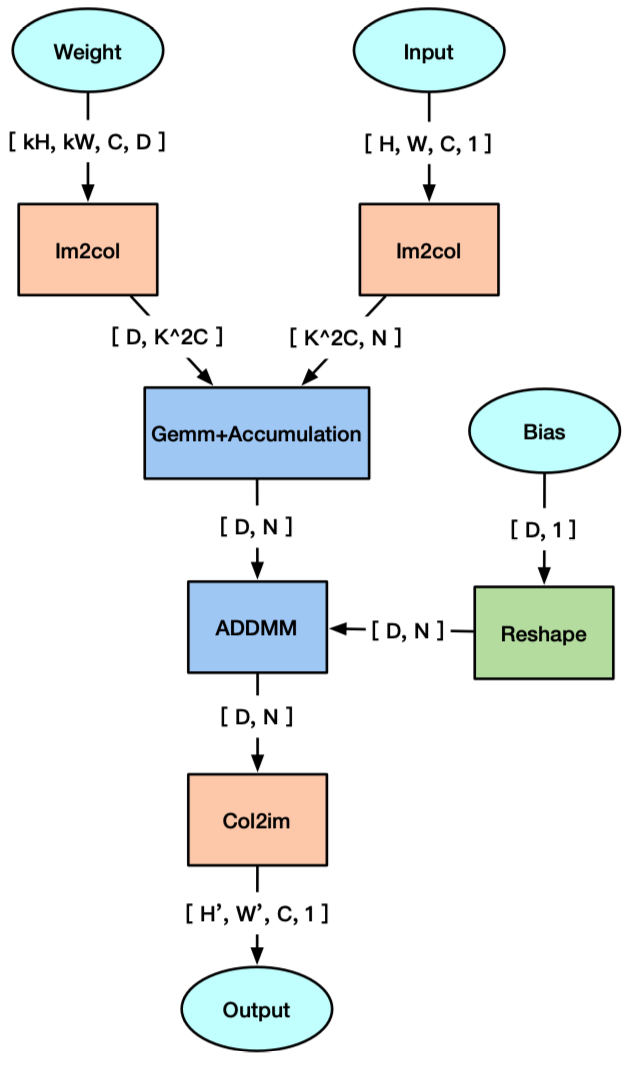}
		\caption{The Forward Graph for convolution computation utilized in most deep learning toolkits like {\small PyTorch}.}
		\label{Figure 2}
	\end{figure}
	
	For the weight $W$, it manually skips the $im2col$ operation and is stored in a bitwise matrix. It is encoded in the direction of rows and each $w_{i,j}$ only takes 1 bit here. The dimention of this bitwise matrix is {\small $[D, K^2C/32]$}.
	
	For the input $X$, it has to firstly pass the $im2col$ operation and gets reshaped into {\small $[K^2C, N]$}. Then, it is encoded in the direction of columns and also stored in a bitwise matrix. Each $x_{i,j}$ only takes 1 bit. The dimention of this bitwise matrix is {\small $[K^2C/32, N]$}.
	
	In {\small PyTorch}, we use {\small $IntTensor$} (int-32) for the weight and {\small $FloatTensor$} (float-32) for the input and the output. 
	
	In our kernel, which is written in {\small $C$}, we use the datatype $uint32\_ t$ for the bitwise weight matrix, the bitwise input matrix and the output matrix. 
	
\begin{figure}[htbp]
		\centering
		\includegraphics[height=10.56cm,width=6cm]{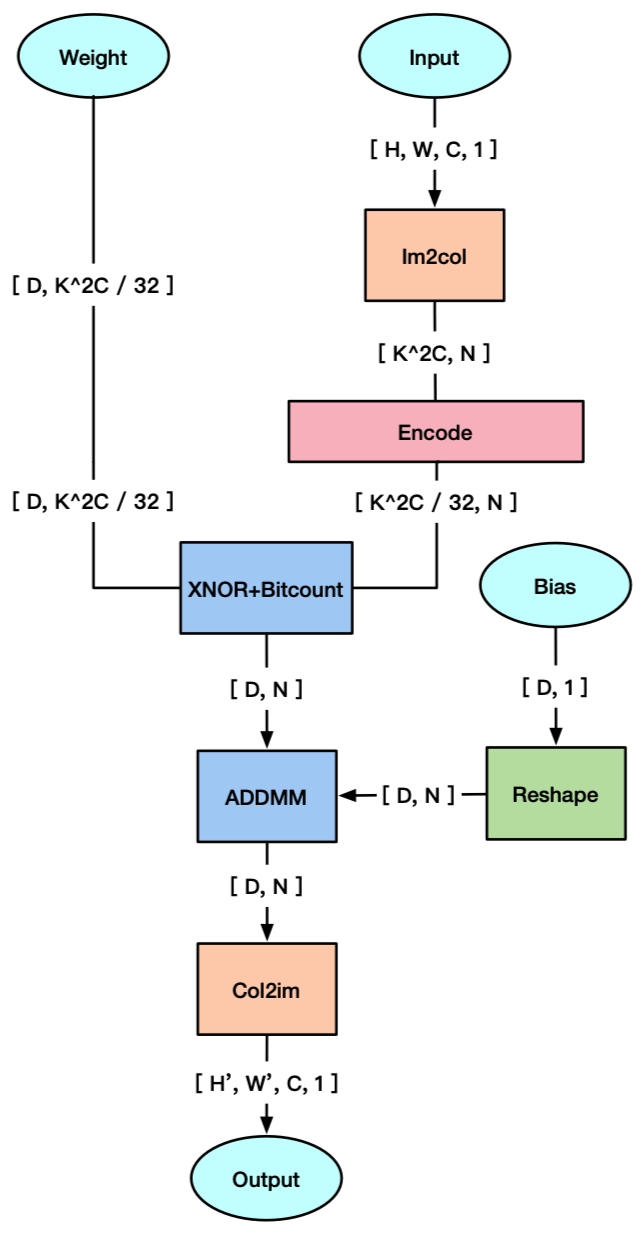}
		\caption{The Forward Graph for convolution computation utilized in our kernel. The input matrix has to be encoded into two bitwise matrixes and then with the encoded weight matrix, they perform the bitwise {\small $Xnor$}-{\small $Bitcount$} operation.}
		\label{Figure 3}
	\end{figure}

	We are not saying that we are also running a simulation like other available binarized neural networks. We encode the input into 1-bit representation mannually and perform the bitwise {\small $Xnor$}-{\small $Bitcount$} operation with the encoded weight. For each element in the bitwise weight matrix, it represents 32 1-bit weights. For each element in the bitwise input matrix, it represents 32 1-bit inputs. The forward graph in our kernel is shown in Figure \ref{Figure 3}.
	
	Notice that both weights and activations are binarized to the binary values, which are either -1 or +1. These binary values are encoded with a 0 for -1 and a 1 for +1. To make it clear, we refer to the binary values -1 and +1 as binary "values" and their encodings 0 and 1 as binary "encodings".

\subsection{Xnor and Bitcount Operation}
	The reason why binarized neural networks can be accelerated is that they replace the {\small $Gemm$}-{\small $Accumulation$} operation with the {\small $Xnor$}-{\small $Bitcount$} operation, which is more hardware friendly and faster.
	
	For the {\small $Gemm$}-{\small $Accumulation$} operation between the float-32 {\small $[D,K^2C]$} weight matrix and the float-32 {\small $[K^2C,N]$} input matrix, each element in the output matrix is calculated as:
	$$
		a_{i,j} = \sum^{K^2C-1}_{k=0}{w_{i,k}}\cdot{x_{k,j}}
	$$
	
	Using an {\small $Xnor$} operation on the binary encodings is equivalent to performing a multiplication operation on the binary values as seen in the following table:
	
	\begin{table}[!htbp]
		\centering
		\begin{tabular}{|cc|c|}
		\hline
		Encoding (Value) & Encoding (Value) & Xnor (Multiply)\\
		\hline
		0 (-1) & 0 (-1) & 1 (+1)\\
		\hline
		0 (-1) & 1 (+1) & 0 (-1)\\
		\hline
		1 (+1) & 0 (-1) & 0 (-1)\\
		\hline
		1 (+1) & 1 (+1) & 1 (+1)\\
		\hline
		\end{tabular}
		\label{table1}
		\vspace{0.3cm}
		\caption{It shows how the {\small $Xnor$} operation on the binary encodings is equivalent to the multiplication operation on the binary values}
	\end{table}

	Therefore, using an {\small $Xnor$}-{\small $Bitcount$} operation on the binary encodings is equivalent to performing a {\small $Gemm$}-{\small $Accumulation$} operation on the binary values. For the {\small $Xnor$}-{\small $Bitcount$} operation between the {\small $[D, K^2C/32]$} bitwise weight matrix and the {\small $[K^2C/32, N]$} bitwise input matrix, each element in the output matrix is calculated as (since we use {\small $0$} to represent {\small $-1$}, it is different from what is calculated directly using {\small $-1$}):
	$$
		a_{i,j} = \sum^{(K^2C/32)-1}_{k=0}2Bitcount(\overline{w_{i,k} \otimes x_{k,j}})-32
	$$

\subsection{Implementation}
	{\small PyTorch} provides the {\small torch.nn} module to help us create and train the neural network. {\small THNN} is a library that gathers {\small torch.nn}'s {\small C} implementations of neural network modules. It can be used in any application that has a {\small C FFI}. There is also a {\small CUDA} counterpart of {\small THNN} in {\small PyTorch}. It is the {\small THCUNN} library in the {\small torch.cunn} module, which provides a {\small CUDA} implementation for many functions in the {\small torch.nn} module.
	
	The CPU implementation of our kernel for the forward graph is based on {\small THNN\_SpatialConvolutionMM\_updateOutput} in the {\small THNN} library. It is using the CPU tensor called {\small THTensor}. Besides, we add an encoding function and replace the {\small $Gemm$}-{\small $Accumulation$} operation function with the {\small $Xnor$}-{\small $Bitcount$} operation function. We keep the rest of its code unchanged.
	
	The bitwise operator of {\small $Xnor$} in {\small C} is {\small $\sim (a\wedge b)$}. We use the bit population count function for counting the number of one-bits from an external library named {\small libpopcnt.h}.
	
	The GPU implementation of our kernel for the forward graph is based on {\small THNN\_SpatialConvolutionMM\_updateOutput} in the {\small THCUNN} library. It is using the GPU tensor called {\small THCTensor}. Like what we do in the CPU implementation, we add an encoding function and replace the {\small $Gemm$}-{\small $Accumulation$} operation function with the {\small $Xnor$}-{\small $Bitcount$} operation function. The rest of its code stays the same.
	
	Both the encoding function and the {\small $Xnor$}-{\small $Bitcount$} operation function are applied on {\small CUDA}. The bitwise operator of $Xnor$ is also {\small $\sim (a\wedge b)$} while for the bit population count function, we use \_\_popc() in the {\small CUDA Toolkit}.

\subsection{Make an Extention for PyTorch}
	{\small PyTorch} provides a plethora of operations related to neural networks. But sometimes, a more customized operation is needed like you might want to use a novel activation function. To address such cases, {\small PyTorch} provides an easy way of writing the customed {\small C} and {\small CUDA} extensions.
	
	After building the {\small C} and {\small CUDA} kernel, we need to connect the kernel with the {\small Python} backend in the {\small THNN} and {\small THCUNN} library. Then, in order to make it a {\small Python} loadable library, we can use the {\small torch.utils.ffi.create\_extension} function in {\small PyTorch}. \footnote{The tutorial for this is at \url{https://github.com/chrischoy/pytorch-custom-cuda-tutorial}.}

\section{Experiments}
\label{label:e}

\subsection{Dataset}
	CIFAR-10 \cite{krizhevsky2010convolutional} is the dataset used in our experiments. It has totally 60,000 images within 10 classes. The size of each image is 32x32x3. It is divided into a training set with 50,000 images and a testing set with 10,000 images. In our experiments, we only use the testing set to test the speed of inference.

\subsection{Model}
	Binarized Neural Network \cite{courbariaux2016binarized} is the network used in our experiments. It is the first network to quantify both weights and activations to binary numbers. 
	
	It uses Deterministic Binarization or the $Sign(x)$ function to binarize parameters, which is quite simple and fast. $Htanh(x)$ is the activation function in the Binarized Neural Network to deal with the gradient mismatch problem \cite{lin2016overcoming}. In backward propagation, gradients are not binary numbers and both weights and activations are updated with real-valued gradients.
	
	Netscope is a tool to visualize neural network architectures. The visualized structure of the Binarized Neural Network used in our experiments can be found \href{http://ethereon.github.io/netscope/#/gist/20cb5ef30b0c43a2f33c5d9625354b16}{here}.

\subsection{Control Group}
	{\small PyTorch} has highly optimized implementations of its operations for CPU and GPU, powered by libraries such as {\small NVIDIA cuDNN} or {\small Intel MKL}. These libraries do not optimize the encoding operation and the  {\small $Xnor$}-{\small $Bitcount$} operation so we do not use them in our computing kernel.
	
	We believe that it is unfair to compare our computing kernel directly with the {\small PyTorch} computing kernel highly optimized by those libraries so we introduce a control group for a fairer comparision.
	
	Like our computing kernel, the kernel in the control group does not have any functions from {\small NVIDIA cuDNN} or {\small Intel MKL}, but it follows the forward graph used in {\small PyTorch}, as shown in Figure \ref{Figure 2}. It is a float-32 computing kernel so it does not encode the parameters to 1 bit and it performs the normal {\small $Gemm$}-{\small $Accumulation$} operation between the weight matrix and the input matrix.

\subsection{Results}
	We equip the Binarized Neural Network with our computing kernel. The computing kernel is only for convolution computation in the model since convolution computation takes up most of the time. We run our test on the inference of the model fed with the CIFAR-10 testing dataset. Results are shown in Table \ref{table2}.
	
	\begin{table}[!htbp]
		\centering
		\begin{tabular}{|c|cc|}
		\hline
		& CPU & GPU\\
		\hline
		{\small PyTorch} & 301s & 1.70s\\
		\hline
		{\small Our Kernel} & 243s & 3.57s\\
		\hline
		{\small Control Group} & 1093s & 11.23s\\
		\hline
		\end{tabular}
		\label{table1}
		\vspace{0.3cm}
		\caption{Results}
	\end{table}
	
	Our kernel is faster in the CPU inference of the Binarized Neural Network than both {\small PyTorch} and {\small Control Group}. It accelerates by about 4.5 times than the kernel in Control Group. Our kernel is faster than the kernel in the {\small Control Group} by about 3 times. The reason why our kernel is slower than the GPU computing kernel in {\small PyTorch} is that the later is highly optimized by {\small NVIDIA cuDNN} but ours not. The specifications of our testing environment is illustrated in the following Table \ref{table2}.
	
	\begin{table}[!htbp]
		\centering
		\begin{tabular}{|c|c|}
			\hline
			CPU & Intel(R) Xeon(R) CPU E5-2620 v4 @ 2.10GHz\\
			\hline
			GPU & NVIDIA Geforce GTX 1080 Ti\\
			\hline
			RAM & 11 GB\\
			\hline
		\end{tabular}
		\label{table2}
		\vspace{0.3cm}
		\caption{The Testing Environment}
	\end{table}

\section{Conclusion}
\label{label:c}
	In this work, we develop a computing kernel for network binarization on {\small PyTorch}. We make a {\small C} and {\small CUDA} extension that supports the encoding operation and the {\small $Xnor$}-{\small $Bitcount$} operation. Experimental results show that our kernel can accelerate the inference of the Binarized Neural Network by about 3 times in GPU and by about 4.5 times in CPU fed with CIFAR-10 testing dataset.

\section{Discussion}
\label{label:d}
	It is known that the bitwise {\small $Xnor$}-{\small $Bitcount$} operation is much simpler than the float-point {\small $Gemm$}-{\small $Accumulation$} operation. This can lead to faster execution time. However, theorizing efficiency speedups is not always precise. For example, we have seen some papers like \cite{rastegari2016xnor} using the total number of instructions in operations as a measure of execution time. We know that the 64-bit x86 instruction set allows a CPU to perform a bitwise {\small $Xnor$} operation within a single instruction while a float-32 {\small $Gemm$} operation takes 32 instructions. But it is not safe to conclude that the {\small $Xnor$} operations would have a 32x speed up over the {\small $Gemm$} operations because instruction and resource scheduling within a CPU is dynamic. We believe that instead of using the total number of instructions in operations as a measure of efficiency speed-ups in network binarization, it is better to look at the actual execution time.
	
	As far as we know, the only available work testing the speed-ups in practice and releasing the code as well is from \cite{courbariaux2016binarized}. They observe a 23x speed up in GPU when binarizing the model. But they are using {\small Theano} \cite{bergstra2010theano} which is not highly optimized in GPU computation as {\small PyTorch}. They are also questioned by some people who try to reproduce their result in the speed up.
	
	Our computing kernel is built and tested on {\small PyTorch}. Our kernel can be accelerated greatly both in CPU and GPU compared with the control group. It means that the bitwise operations could help binarized neural networks achieve a speed up in execution time. Although our kernel is a little bit faster in the CPU test than the CPU computing kernel in {\small PyTorch}, it is much slower than the GPU computing kernel in it. The reason is that the computing kernel in {\small PyTorch} is highly optimized by libraries like {\small NVIDIA cuDNN} or {\small Intel MKL}, especially the GPU computing kernel. Therefore, for the inference of binarized neural network on CPU, our kernel is faster, but on GPU, running the simulation on {\small PyTorch} seems a better idea unless {\small NVIDIA cuDNN} optimizes the 1-bit convolution computation in the future.

\section{Acknowledgement}
	We would like to thank Mitacs Globalink Program for its funding to support Xianda to conduct the research internship.

\newpage


\end{document}